\documentclass[letterpaper]{article}
\usepackage{arxiv-preprint}
\usepackage[hyphens]{url}
\usepackage{graphicx}
\usepackage{natbib}
\usepackage{caption}
\usepackage{algorithm}
\usepackage{algorithmic}
\usepackage{newfloat}
\usepackage{listings}
\DeclareCaptionStyle{ruled}{labelfont=normalfont,labelsep=colon,strut=off}
\floatstyle{ruled}
\newfloat{listing}{tb}{lst}{}
\floatname{listing}{Listing}
\usepackage{booktabs}
\usepackage{amsmath}
\usepackage{amssymb}
\usepackage{multirow}

\newcommand{\ourUA}{78.90$\pm$2.80}
\newcommand{\ourWA}{78.18$\pm$1.44}
\newcommand{\caseACC}{61.90}
\newcommand{\caseWF}{59.82}
\newcommand{\ourMELDWA}{65.95}
\newcommand{\ourMELD}{64.33}

\title{BiCFlow-MER: Orchestrating Discriminative and Generative Multimodal Emotion Recognition via Conditional Transport}
\author{
    Yanbing Wang\equalcontrib,
    Shenyue Wang\equalcontrib,
    Chunyang Yu\corresponding
}
\affiliations{
    OPPO Research Institute\\
    \{wangyanbing1, wangshenyue, yuchunyang\}@oppo.com
}

\begin{document}

\nocopyright
\maketitle

\begin{abstract}
In multimodal emotion recognition (MER), human affective states are inferred by integrating complementary cues from multiple modalities.
In audio-text MER, affective cues are often entangled with speaker style and lexical content, while cross-modal disagreement further complicates how the evidence should be integrated.
Under conventional discriminative fusion, multimodal evidence is compressed into a terminal prediction, with modality-specific cues and conflict information insufficiently preserved.
In large generative affective models, by contrast, affective reasoning is typically embedded in language decoding, leaving emotion evidence implicit and difficult to verify in a structured space.
To address these limitations, BiCFlow-MER (Bidirectional Conditional Flow for Multimodal Emotion Recognition) is proposed as a conditional-flow framework in which audio-text MER is formulated as generative evidence transport within a structured emotion space.
Within BiCFlow-MER, emotion-oriented evidence is disentangled from speaker-style and lexical-content factors to construct a conflict-aware affective condition.
Guided by this condition, each utterance is transported to an explicit emotion-space endpoint through a bidirectional rectified flow.
Candidate emotions are jointly scored through adaptive prototype-cloud geometry of the transported endpoint and backward compatibility with the condition-dependent start region, enabling conflict-aware recognition.
BiCFlow-MER achieves the best reported results among all compared methods across IEMOCAP, MELD, and the zero-shot CASE benchmark.
By orchestrating discriminative recognition and generative evidence modeling through conditional transport, BiCFlow-MER defines a new MER paradigm.
\end{abstract}
\section{Introduction}

MER aims to infer a speaker's affective state from behavioral signals such as speech, transcripts, and visual cues~\cite{heredia2022adaptive,pena2023framework}.
Beyond its scientific role in affective computing, MER supports customer-experience analytics, human-computer interaction, intelligent assistants, and mental-health applications.
Industry demand reflects this direction: a MarketsandMarkets report estimates the global emotion detection and recognition market at \$26.76B in 2025 and \$43.29B by 2031, with an 8.35\% CAGR from 2024-2031.
The report also identifies customer-experience analytics as the largest 2031 application segment and multimodal data fusion as the fastest-growing technology direction.
These trends create a practical requirement for MER: emotion must be recognized reliably from signals already available in real-world interfaces, without introducing substantial sensing cost or privacy burden.
Audio and transcripts meet this requirement better than visual streams, which often depend on camera availability, bandwidth, lighting, and device placement~\cite{ghosh2022mmer,lin2023robust,wang2025enhancing}.
Yet real-world audio and text do not provide clean or consistently aligned emotion evidence: vocal affect is entangled with speaker style, lexical affect with semantic content, and tone may contradict words.
Most audio-text MER systems nevertheless fuse these signals and map the fused representation directly to an emotion label~\cite{tsai2019multimodal,wang2024inter,fan2025coordination,wang2025enhancing}.
This terminal decision leaves three evidence-level questions only indirectly modeled: which part of the signal reflects emotion rather than speaker style or lexical content; how audio-text disagreement should modulate fusion; and whether a candidate emotion remains compatible with the observed evidence~\cite{zhang2024camel,liu2026fine,huang2026tone}.

Generative modeling offers one way to make affective evidence formation more explicit.
Diffusion and rectified-flow models connect simple priors to structured data distributions through continuous trajectories~\cite{liu2022flow,peebles2023scalable}, while large language model (LLM) and multimodal large language model (MLLM) affective systems popularize intermediate reasoning and explanatory affect analysis~\cite{xu2024secap,li2026llms,rha2026emotion}.
For audio-text MER, however, large generative affective systems often couple recognition to token decoding, model scale, prompt design, or additional visual and conversational context.
The relevant question is therefore how to obtain a compact generative evidence object that can be checked against the original speech and language signals.

This paper introduces \textbf{BiCFlow-MER}, a conditional flow framework for audio-text MER.
BiCFlow-MER first builds an affective condition $\mathbf{c}$ by separating emotion-oriented evidence from speaker style and lexical content and by using audio-text conflict to modulate fusion.
Conditional Trajectory Bidirectional Rectified Flow (CT-BiRF) then transports a condition-dependent start point toward an emotion-space endpoint $\hat{\mathbf{e}}_1$, making the generated endpoint an explicit evidence object for recognition.
Adaptive Geometric Classification Head (A-GCH) and Bidirectional Generative Decision (BGD) evaluate this endpoint through forward prototype-cloud geometry and backward compatibility scoring.
The resulting pipeline turns audio-text evidence into a transportable endpoint and then verifies the endpoint geometrically, linking evidence construction, conditional flow, and class-level decision checking in a single recognition process. The main contributions are organized at the paradigm, module, and empirical levels:
\begin{itemize}
    \item \textbf{Generative-decision paradigm.} BiCFlow-MER replaces terminal fusion-to-label mapping with conditional transport to an explicit emotion-space endpoint followed by forward geometric scoring and backward compatibility scoring.
    \item \textbf{Evidence construction and scoring modules.} Emotion-oriented factorization and ECA construct a conflict-aware affective condition, while CT-BiRF, A-GCH, and BGD perform endpoint transport and bidirectional class scoring.
    \item \textbf{Empirical validation.} Experiments on IEMOCAP, MELD, and zero-shot CASE, together with controlled variants, evaluate standard recognition, conflict transfer, and the roles of the main components.
\end{itemize}

\section{Related Work}

\paragraph{Discriminative audio-text MER.}
Most utterance-level MER systems encode speech and language with pretrained backbones, align or attend across modalities, and map the fused representation to emotion labels~\cite{tsai2019multimodal,ghosh2022mmer,wang2025enhancing}.
Recent variants improve this pipeline with modality-specific self-supervision, empirical fusion design, cross-space synergy, incomplete-modality prompting, and consistency-aware learning~\cite{yu2021learning,wu2023empirical,lyu2026cross,he2026cross,yin2026tical}.
These methods improve representation learning but retain terminal classification, leaving evidence formation and class compatibility implicit. BiCFlow-MER instead transports and geometrically verifies the fused evidence.

\paragraph{Affective evidence factorization.}
Another line separates emotion from speaker style, lexical content, or other nuisance factors through factorized subspaces, variational bottlenecks, and Contrastive Log-ratio Upper Bound (CLUB)-based mutual-information control~\cite{liu2026fine,kingma2013auto,cheng2020club}.
Factorization improves affective representations but stops before post-fusion evidence organization and verification. BiCFlow-MER uses emotion-oriented factors to construct the transport condition.

\paragraph{Generative modeling and conditional transport.}
Generative affective systems, including LLM and MLLM-based methods, have introduced richer intermediate structures such as emotional descriptions and rationale-based affect analysis~\cite{wu2025enriching,li2026llms,rha2026emotion}.
Continuous generative models provide another route: rectified flow learns transport trajectories from simple priors to structured data distributions~\cite{liu2022flow}, and transformer velocity fields extend this formulation~\cite{peebles2023scalable}.
Recent speech-emotion and ambiguity-aware methods also explore distributional or transport-inspired supervision~\cite{wang2026gen,wu2026amber}.
These methods produce textual explanations or distributional supervision but do not construct and verify an audio-text emotion endpoint. BiCFlow-MER generates a condition-dependent endpoint.

\paragraph{Prototype geometry and verification.}
Distance- and prototype-based decisions are well established in metric learning~\cite{schroff2015facenet}.
Prototype scoring measures class proximity but not compatibility with the condition-dependent start region. BiCFlow-MER combines adaptive prototype clouds with backward class-to-condition compatibility scoring.

\section{Method}

\subsection{Overview and Problem Formulation}
\begin{figure*}[t]
\centering
\includegraphics[width=\textwidth]{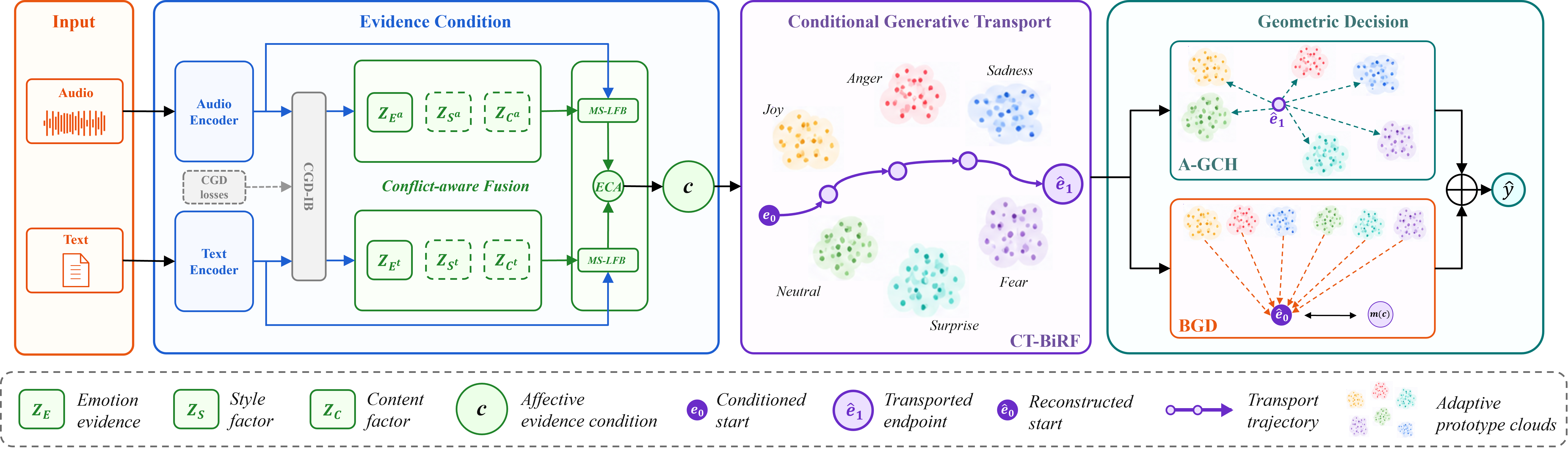}
\caption{BiCFlow-MER pipeline. Encoder features and CGD-IB emotion factors are combined through multi-scale fusion and ECA to form the affective condition $\mathbf{c}$. CT-BiRF transports a condition-dependent start $\mathbf{e}_0=m(\mathbf{c})+\boldsymbol{\sigma}_0\odot\boldsymbol{\varepsilon}_0$ to the endpoint $\hat{\mathbf{e}}_1$, which A-GCH scores against adaptive prototype clouds. BGD maps each class prototype mean $\boldsymbol{\mu}_k$ back to a reconstructed start $\hat{\mathbf{e}}_0^{(k)}$ and compares it with the prior center $m(\mathbf{c})$. A single reconstruction and the displayed class clouds are schematic.}
\label{fig:pipeline}
\end{figure*}

Given paired utterance-level audio $\mathbf{x}_a$ and transcript $\mathbf{x}_t$, the goal is to predict emotion label $y \in \{1,\ldots,K\}$.
BiCFlow-MER organizes recognition as evidence construction, conditional transport, and bidirectional geometric scoring.
It maps the inputs to an affective condition $\mathbf{c} \in \mathbb{R}^{d_c}$, transports a condition-dependent start through a $d$-dimensional emotion flow space $\mathcal{E}$ ($d{=}64$), and evaluates the generated endpoint $\hat{\mathbf{e}}_1 \in \mathcal{E}$ in both forward and backward directions:
\begin{equation}
\begin{aligned}
(\mathbf{x}_a, \mathbf{x}_t) &\xrightarrow{\text{evidence}} \mathbf{c}
\xrightarrow{\text{transport}} \hat{\mathbf{e}}_1,\\
(\hat{\mathbf{e}}_1,\mathbf{c})
&\xrightarrow{\text{geometric scoring}}
\{\ell_k^{\mathrm{fwd}},\ell_k^{\mathrm{bwd}}\}_{k=1}^{K}
\longrightarrow \hat{y}.
\end{aligned}
\end{equation}
Figure~\ref{fig:pipeline} summarizes the architecture.
For modality $m\in\{a,t\}$, $\mathbf{z}_E^m$, $\mathbf{z}_S^m$, and $\mathbf{z}_C^m$ denote emotion, style, and content factors; $\mathbf{e}_t$ is the flow state at $t\in[0,1]$; and $(\boldsymbol{\mu}_k,\boldsymbol{\Sigma}_k)$ parameterizes the adaptive prototype cloud of class $k$.
The forward score tests whether $\hat{\mathbf{e}}_1$ belongs to a class cloud, while the backward score tests whether that class hypothesis reconstructs the condition-dependent start region.

\subsection{Affective Evidence Factorization}
BiCFlow-MER first separates emotion-oriented evidence from style and content variation in each modality.
In the IEMOCAP/MELD configuration, audio and text are encoded by WavLM-Large and DeBERTa-v3 with low-rank adaptation (LoRA) fine-tuning~\cite{chen2022wavlm,he2020deberta,hu2022lora}:
\begin{equation}
\mathbf{h}_m = f_m(\mathbf{x}_m), \quad m \in \{a,t\}.
\end{equation}
The Causal Generative Disentanglement with an Information Bottleneck (CGD-IB) block mask-pools each encoded sequence and uses three $d_z$-dimensional variational heads for $r\in\{E,S,C\}$:
\begin{equation}
\begin{aligned}
q_\phi(\mathbf{z}_r^m\mid\mathbf{h}_m)
&=\mathcal{N}\!\left(\boldsymbol{\mu}_r^m,
\operatorname{diag}((\boldsymbol{\sigma}_r^m)^2)\right),\\
\mathbf{z}_r^m
&=\boldsymbol{\mu}_r^m+
\boldsymbol{\sigma}_r^m\odot\boldsymbol{\varepsilon}_r^m,
\quad \boldsymbol{\varepsilon}_r^m\sim
\mathcal{N}(\mathbf{0},\mathbf{I}_{d_z}).
\end{aligned}
\end{equation}
Table~\ref{tab:factor_roles} summarizes the contract imposed on these factors.

\begin{table}[t]
\centering
\small
\begin{tabular*}{\columnwidth}{@{\extracolsep{\fill}}lll@{}}
\toprule
Factor & Role & Routed to \\
\midrule
$\mathbf{z}_E^m$ & affective evidence & condition, prototype anchor \\
$\mathbf{z}_S^m$ & speaker/style variation & reconstruction, regularization \\
$\mathbf{z}_C^m$ & lexical/content variation & reconstruction, regularization \\
\bottomrule
\end{tabular*}
\caption{Roles of the modality-specific factors produced by CGD-IB.}
\label{tab:factor_roles}
\end{table}

The active factorization objective is
\begin{equation}
\mathcal{L}_{\mathrm{CGD}}
=\mathcal{L}_{\mathrm{rec}}^{a}
+\mathcal{L}_{\mathrm{rec}}^{t}
+\beta_{\mathrm{IB}}\mathcal{L}_{\mathrm{KL}}
+\lambda_{\mathrm{cf}}\mathcal{L}_{\mathrm{cf}}
+\lambda_E\mathcal{L}_{E}.
\end{equation}
$\mathcal{L}_{\mathrm{rec}}^{a}$ and $\mathcal{L}_{\mathrm{rec}}^{t}$ reconstruct the representations supplied to their respective audio/acoustic and text/semantic branches from the three branch factors, and $\mathcal{L}_{\mathrm{KL}}$ regularizes all variational posteriors.
For $\mathcal{L}_{\mathrm{cf}}$, style factors are exchanged across different-label minibatch samples, the resulting features are decoded, and their emotion prediction is constrained to retain the source label.
$\mathcal{L}_{E}$ directly constrains each emotion factor to preserve label-relevant semantics during factorization.

The factor-level classifiers optimized by $\mathcal{L}_{E}$ produce modality-specific class probabilities from $\mathbf{z}_E^a$ and $\mathbf{z}_E^t$.
Let $p_m^{\max}$ be the maximum class probability of the factor-level classifier for modality $m$.
From Stage~II.a onward, modality dropout~\cite{wu2023empirical} removes the corresponding factor with probability
\begin{equation}
p_{\mathrm{drop}}^m=\frac{p_0}{2}\,p_m^{\max}.
\end{equation}
With base dropout rate $p_0{=}0.2$, at most one modality is dropped per sample, and higher $p_m^{\max}$ increases its masking probability.
Only $\mathbf{z}_E^m$ is routed to downstream condition construction; the dashed $\mathbf{z}_S^m$ and $\mathbf{z}_C^m$ factors in Figure~\ref{fig:pipeline} remain within CGD-IB for reconstruction and regularization.

\subsection{Conflict-Aware Condition Construction}
The encoded sequences and the two emotion factors are then combined into the transport condition $\mathbf{c}$.
\paragraph{Multi-scale fusion.}
A multi-scale latent fusion bottleneck (MS-LFB) projects the audio and text encoder sequences into a shared space and concatenates them along the temporal dimension.
Three parallel paths extract global summaries through learned-query attention, local salient evidence through Gumbel Top-$K$ selection, and short-term temporal patterns through one-dimensional convolutions.
The shared MS-LFB is evaluated under two complementary modality masks.
For the audio-specific pass, the text mask and $\mathbf{z}_E^t$ are set to zero; for the text-specific pass, the audio mask and $\mathbf{z}_E^a$ are set to zero.
Each multi-scale summary is concatenated with its retained emotion factor and projected by the fusion MLP, producing $\mathbf{c}_a^{\mathrm{fuse}},\mathbf{c}_t^{\mathrm{fuse}}\in\mathbb{R}^{d_f}$.
These two masked passes correspond to the two MS-LFB branches in Figure~\ref{fig:pipeline}.

\paragraph{Conflict-conditioned fusion.}
The Emotion Conflict Awareness (ECA) module assigns an energy to each audio-text emotion-factor pair:
\begin{equation}
\label{eq:eca_energy}
\begin{aligned}
E_\psi
={}&\bigl\|p_\psi(\mathbf{z}_E^a)-p_\psi(\mathbf{z}_E^t)\bigr\|_2^2\\
&+\lambda_{\cos}\bigl(1-\operatorname{cos}(\mathbf{z}_E^a,\mathbf{z}_E^t)\bigr)\\
&+s_\psi([\mathbf{z}_E^a;\mathbf{z}_E^t]).
\end{aligned}
\end{equation}
Here, $p_\psi$ is a shared projector, $s_\psi$ is a learned scalar pair scorer, $\lambda_{\cos}{=}0.5$ weights cosine disagreement, and $\gamma{=}0.1$ scales the standardized conflict-energy residual in the fused condition.
After standardization to $\widetilde{E}_\psi$ using running statistics accumulated during training and frozen at inference, ECA predicts conflict-conditioned modality weights and constructs
\begin{equation}
\label{eq:eca_condition}
\begin{aligned}
\mathbf{q}
&=g_\omega([\mathbf{z}_E^a;\mathbf{z}_E^t;\widetilde{E}_\psi])
\in\mathbb{R}^{2},\\
(\alpha_a,\alpha_t)&=\operatorname{softmax}(\mathbf{q}),\\
\bar{\mathbf{c}}
&=\alpha_a\mathbf{c}_a^{\mathrm{fuse}}
+\alpha_t\mathbf{c}_t^{\mathrm{fuse}}
+\gamma\widetilde{E}_\psi\,\mathbf{1}_{d_f},\\
\mathbf{c}&=\mathbf{W}_c\bar{\mathbf{c}}+\mathbf{b}_c.
\end{aligned}
\end{equation}
$g_\omega$ is a learned modality-weight scorer optimized end-to-end with the downstream objectives.
ECA uses the hinge-ranking objective $\mathcal{L}_{\mathrm{ECA}}=\sum_{j\in\mathcal{A}}w_j[\delta+E_\psi^+-E_{\psi,j}^-]_+$ with margin $\delta{=}0.5$, weights $(w_1,w_2,w_3){=}(0.3,0.4,0.3)$, overall weight 0.1, and a Huber energy regularizer weighted by 0.001.
Here, $E_\psi^+$ is the matched-pair energy; N1 (cross-emotion mismatches) and N2 (same-emotion noncorresponding pairs) form $\mathcal{A}{=}\{1,2\}$ in Stage~II.a, while Stage~II.b uses $\mathcal{A}{=}\{1,2,3\}$ by adding N3 (independently shuffled pairs).

\subsection{CT-BiRF}
CT-BiRF generates a compact endpoint in the emotion flow space $\mathcal{E}$ from the affective condition $\mathbf{c}$.
Building on rectified flow~\cite{liu2022flow}, it uses forward and backward velocity fields $\mathbf{v}_f$ and $\mathbf{v}_b$ with a shared Diffusion Transformer (DiT)-style trunk and direction-specific blocks and output heads.
The affective condition $\mathbf{c}$ parameterizes both fields throughout forward transport and backward compatibility scoring.
The two fields model opposite directions along the same conditional path:
\begin{equation}
\begin{aligned}
\mathbf{v}_f(\mathbf{e}_t,t,\mathbf{c})
&\approx\mathbf{e}_1-\mathbf{e}_0,\\
\mathbf{v}_b(\mathbf{e}_t,t,\mathbf{c})
&\approx\mathbf{e}_0-\mathbf{e}_1.
\end{aligned}
\end{equation}

\paragraph{Conditional prior.}
Instead of sampling $\mathbf{e}_0 \sim \mathcal{N}(\mathbf{0}, \mathbf{I}_d)$, the start point is drawn as
\begin{equation}
\mathbf{e}_0=m(\mathbf{c})+
\boldsymbol{\sigma}_0\odot\boldsymbol{\varepsilon}_0,
\quad
\boldsymbol{\varepsilon}_0\sim\mathcal{N}(\mathbf{0},\mathbf{I}_d),
\end{equation}
where an MLP predicts the deterministic center $m(\mathbf{c})$ and $\boldsymbol{\sigma}_0$ is a learned diagonal scale.
Thus, $m(\mathbf{c})$ identifies the condition-dependent start region, while $\mathbf{e}_0$ is a stochastic point sampled from that region.

\paragraph{Terminal targets and prototype anchoring.}
The target distribution is estimated from emotion factors rather than generated endpoints.
For each minibatch, the post-dropout modality factors are averaged and projected into $\mathcal{E}$:
\begin{equation}
\mathbf{r}=\mathbf{W}_p\bar{\mathbf{z}}_E,
\qquad
\bar{\mathbf{z}}_E=\tfrac{1}{2}(\mathbf{z}_E^a+\mathbf{z}_E^t).
\end{equation}
Detached vectors $\mathbf{r}$ are stored in a per-class first-in-first-out (FIFO) memory bank $\mathcal{M}_k$ with capacity 256.
The bank estimates a mean $\boldsymbol{\mu}_k$ and an adaptive diagonal-plus-low-rank covariance
\begin{equation}
\boldsymbol{\Sigma}_k
=\mathbf{D}_k+\mathbf{U}_k\mathbf{U}_k^\top
+\alpha_k\bar{v}\,\mathbf{I}_d,
\end{equation}
where $\mathbf{D}_k$ is the diagonal covariance component, $\mathbf{U}_k\mathbf{U}_k^\top$ captures rank-4 correlations, $\mathbf{I}_d$ is the $d$-dimensional identity matrix, and $\bar{v}$ is the global mean within-class variance.
For class count $n_k$, shrinkage is $\alpha_k=\min\{0.1/\sqrt{n_k/100},0.5\}$.
The terminal target for a sample with label $y$ is then drawn from the resulting prototype cloud:
\begin{equation}
\mathbf{e}_1\sim\mathcal{N}(\boldsymbol{\mu}_y,\boldsymbol{\Sigma}_y).
\end{equation}
This construction lets the same class clouds define both flow targets and geometric decisions without feeding $\hat{\mathbf{e}}_1$ back into their statistics.

\paragraph{Class-conditioned stochastic interpolation.}
During training, interpolated states are perturbed:
\begin{equation}
\begin{aligned}
\mathbf{e}_t
&=(1-t)\mathbf{e}_0+t\mathbf{e}_1
+\boldsymbol{\sigma}(\mathbf{c},t,y)\odot\boldsymbol{\varepsilon}_t,\\
\mathbf{0}&\preceq\boldsymbol{\sigma}(\mathbf{c},t,y)
\preceq\sigma_{\max}\mathbf{1}_d.
\end{aligned}
\end{equation}
Here, $\boldsymbol{\varepsilon}_t\sim\mathcal{N}(\mathbf{0},\mathbf{I}_d)$, $\sigma_{\max}{=}0.1$, and $\preceq$ denotes an elementwise bound on the scale predicted from the condition, time, and label.
The perturbation is used as off-path neighborhood augmentation, while the supervised velocity remains the nominal rectified displacement $\mathbf{e}_1-\mathbf{e}_0$.

\paragraph{Bidirectional flow losses.}
Let $\boldsymbol{\Delta}=\mathbf{e}_1-\mathbf{e}_0$.
The flow-matching and bidirectional consistency losses are
\begin{equation}
\begin{aligned}
\mathcal{L}_{\mathrm{rf}}^f
&=\mathbb{E}\!\left[\kappa_y
\|\mathbf{v}_f(\mathbf{e}_t,t,\mathbf{c})-\boldsymbol{\Delta}\|_2^2\right],\\
\mathcal{L}_{\mathrm{rf}}^b
&=\mathbb{E}\!\left[
\|\mathbf{v}_b(\mathbf{e}_t,t,\mathbf{c})+\boldsymbol{\Delta}\|_2^2\right],\\
\mathcal{L}_{\mathrm{bc}}
&=\mathbb{E}\!\left[
\|\mathbf{v}_f(\mathbf{e}_t,t,\mathbf{c})
+\mathbf{v}_b(\mathbf{e}_t,t,\mathbf{c})\|_2^2\right],
\end{aligned}
\end{equation}
where $\kappa_y=1+\beta_\kappa\operatorname{tr}(\boldsymbol{\Sigma}_y)\big/\left(K^{-1}\sum_{k=1}^{K}\operatorname{tr}(\boldsymbol{\Sigma}_k)\right)$ weights forward supervision by the normalized trace of the target-class prototype covariance, with $\beta_\kappa{=}0.5$.
Both fields are evaluated at the same interpolated state and time, enforcing opposite velocities along the same conditional path.
Terminal prototype inversion is evaluated at $t=1$.
For two independently sampled times on the same conditional path, sharing $(\mathbf{e}_0,\mathbf{e}_1,\mathbf{c})$ while using independent stochastic perturbations, define
\begin{equation}
\begin{aligned}
\mathbf{f}_{t_i}
&=\mathbf{e}_{t_i}+(1-t_i)
\mathbf{v}_f(\mathbf{e}_{t_i},t_i,\mathbf{c}),\\
\mathcal{L}_{\mathrm{ct}}
&=\mathbb{E}\|\mathbf{f}_{t_1}-\mathbf{f}_{t_2}\|_2^2.
\end{aligned}
\end{equation}
This term aligns extrapolated endpoints; relative to unit-weight forward matching, backward matching, bidirectional consistency, and trajectory consistency are each weighted by 0.5.
Stage~II.b further applies the backward landmark loss
\begin{equation}
\mathcal{L}_\mathrm{back}
=\left\|\mathbf{e}_1+\mathbf{v}_b(\mathbf{e}_1,1,\mathbf{c})
-m(\mathbf{c})\right\|_2^2,
\end{equation}
which trains terminal points to invert toward the deterministic center of the condition-dependent start region.

\paragraph{Inference.}
The primary endpoint is generated by one Euler step:
\begin{equation}
\hat{\mathbf{e}}_1
=\mathbf{e}_0+\mathbf{v}_f(\mathbf{e}_0,0,\mathbf{c}).
\end{equation}
For $M{=}8$ stochastic starts, the resulting class logits are aggregated by log-mean-exp:
\begin{equation}
\bar{\ell}_k
=\log\!\left(\frac{1}{M}\sum_{j=1}^{M}
\exp(\ell_k^{(j)})\right).
\end{equation}
Samples with maximum A-GCH probability below 0.7 replace the one-step endpoint with a three-step Euler integration before geometric rescoring.

\subsection{Geometric Evidence and Backward Compatibility}
\paragraph{A-GCH.}
The Adaptive Geometric Classification Head (A-GCH) evaluates $\hat{\mathbf{e}}_1$ under each Gaussian prototype cloud.
Its raw Gaussian score contains both the squared Mahalanobis distance and the covariance normalization term:
\begin{equation}
g_k
=-\frac{1}{2}d_M^2\!\left(
\hat{\mathbf{e}}_1;\boldsymbol{\mu}_k,\boldsymbol{\Sigma}_k\right)
-\frac{1}{2}\log\det(\boldsymbol{\Sigma}_k).
\end{equation}
At inference, conflict-adaptive temperature scaling gives
\begin{equation}
\ell_k^{\mathrm{fwd}}=\frac{g_k}{\tau(\mathbf{c})},
\qquad
\tau(\mathbf{c})
=\tau_0\bigl(1+\eta\max\{0,\widetilde{E}_\psi\}\bigr).
\end{equation}
The transported endpoint is directly supervised by
\begin{equation}
\mathcal{L}_{\mathrm{cls}}^{\mathrm{flow}}
=\mathrm{CE}\bigl(\{g_k\}_{k=1}^{K},y\bigr),
\end{equation}
so the flow field learns endpoints that are likely under the correct adaptive cloud.

\paragraph{BGD.}
Forward scoring identifies the prototype cloud that best explains $\hat{\mathbf{e}}_1$.
Bidirectional Generative Decision (BGD) provides a backward compatibility score indicating whether the mean of a candidate class can invert toward the condition-dependent start region.
For each class $k$,
\begin{equation}
\hat{\mathbf{e}}_0^{(k)} = \boldsymbol{\mu}_k + \mathbf{v}_b(\boldsymbol{\mu}_k, 1, \mathbf{c}),
\end{equation}
and its consistency with the conditional-prior center is scored by
\begin{equation}
\ell_k^{\mathrm{bwd}}
=-\frac{\left\|\hat{\mathbf{e}}_0^{(k)}-m(\mathbf{c})\right\|_2^2}
{2\tau_{\mathrm{back}}}.
\end{equation}
Thus, the $m(\mathbf{c})$ marker in Figure~\ref{fig:pipeline} is the deterministic reference shared by the stochastic forward starts and the class-specific backward reconstructions.
Accordingly, $\hat{\mathbf{e}}_1$ supplies the forward A-GCH branch, whereas each pair $(\boldsymbol{\mu}_k,\mathbf{c})$ supplies the backward BGD branch.
BGD introduces no additional trainable parameters or class-wise objective; it reuses the learned backward field to construct this inference-time compatibility score.
Final logits combine both channels in log space:
\begin{equation}
\ell_k = \ell_k^{\mathrm{fwd}} + \lambda_{\mathrm{back}}\, \ell_k^{\mathrm{bwd}},
\end{equation}
with $\lambda_\mathrm{back}{=}0.5$ and temperature $\tau_\mathrm{back}{=}2$ selected on validation.
The final score favors classes whose cloud explains the transported endpoint and whose mean reconstructs the original condition-dependent region.

\subsection{Training Objective and Curriculum}
Training activates evidence construction, transport, and inverse-flow refinement in sequence.
Stage~I learns the variational emotion factors using the CGD-IB objective described above.
Stage~II.a retains this factorization supervision and introduces ECA, bidirectional flow matching, trajectory consistency, prototype-cloud classification, and memory-bank updates.
Stage~II.b activates the additional N3 conflict pairs and backward landmark supervision to refine terminal inversion.
Each stage optimizes only the objectives activated at that stage; BGD adds no separate training loss.
Table~\ref{tab:curriculum} summarizes the curriculum.

\begin{table}[t]
\centering
\small
\begin{tabular*}{\columnwidth}{@{\extracolsep{\fill}}cll@{}}
\toprule
Stage & Epochs & Training focus \\
\midrule
I & 5 & Affective evidence factorization \\
II.a & 10 & Conditional transport and geometric scoring \\
II.b & 20 & Conflict arbitration and inverse-flow refinement \\
\bottomrule
\end{tabular*}
\caption{Three-stage training curriculum.}
\label{tab:curriculum}
\end{table}

\section{Experiments}

\begin{table}[!tb]
\centering
\footnotesize
\setlength{\tabcolsep}{2.5pt}
\renewcommand{\arraystretch}{1.08}

\begin{tabular*}{\columnwidth}{
@{}p{0.24\columnwidth}@{\extracolsep{\fill}}cccccc@{}}
\toprule
\multirow{2}{*}{Method}
& \multirow{2}{*}{S}
& \multirow{2}{*}{T}
& \multicolumn{2}{c}{IEMOCAP}
& \multicolumn{2}{c}{MELD} \\
\cmidrule(lr){4-5}
\cmidrule(l){6-7}
& & & WA(\%) & UA(\%) & WA(\%) & W-F1(\%) \\
\midrule

RobinNet
& \checkmark & \checkmark
& 73.00 & 73.00 & 62.00 & 46.29 \\

MER-HAN
& \checkmark & \checkmark
& 73.33 & 74.20 & 62.87 & 60.22 \\

GM2F
& \checkmark & \checkmark
& 75.98 & 77.43 & -- & 61.27 \\

MCFN
& \checkmark & \checkmark
& 76.01 & 77.84 & 64.50 & 62.20 \\

SAMS
& \checkmark & \checkmark
& 76.60 & 78.10 & \underline{65.40} & 62.60 \\

LSGMER
& \checkmark & \checkmark
& 77.50 & 77.90 & -- & \underline{62.90} \\

SCMI-Net
& \checkmark & \checkmark
& 77.60 & 78.60 & 65.00 & 62.30 \\

MIUST
& \checkmark & \checkmark
& \underline{77.78} & \underline{78.79}
& 64.20 & 60.74 \\

\midrule

\textbf{Ours}
& \checkmark & \checkmark
& \textbf{\ourWA}
& \textbf{\ourUA}
& \textbf{\ourMELDWA}
& \textbf{\ourMELD} \\

\bottomrule
\end{tabular*}

\caption{
Audio-text MER comparison on four-class IEMOCAP and
seven-class MELD. BiCFlow-MER uses five-fold speaker-independent
training and evaluation over the five IEMOCAP sessions; reported
IEMOCAP values are the mean $\pm$ standard deviation across the five
held-out-session folds. MELD results follow the
official train/development/test splits on MELD. Best results are
bolded and second-best results are underlined.
}
\label{tab:external_compare}
\end{table}

\subsection{Experimental Setup}
\paragraph{Dataset and protocol.}
Evaluation is conducted on four-class IEMOCAP~\cite{busso2008iemocap}: angry, happy (merged excited/happy), sad, and neutral, under the utterance-level audio-text setting.
We use five-fold speaker-independent evaluation over its five sessions, training on four sessions and testing on the held-out session.
We also evaluate BiCFlow-MER on MELD~\cite{poria2019meld} under the seven-class setting with its official train/development/test splits.
Metrics are unweighted accuracy (UA), weighted accuracy (WA), and weighted F1 (W-F1); MELD results are reported with WA and W-F1, with W-F1 as the main metric.

\paragraph{Modalities.}
For IEMOCAP and MELD, BiCFlow-MER encodes audio with WavLM-Large and text with DeBERTa-v3-base.
Each utterance is classified independently without visual features, speaker-identity metadata, or conversational context; unless stated otherwise, ablations use the same bimodal input.

\paragraph{Implementation details.}
For IEMOCAP and MELD, audio is resampled to 16\,kHz mono and transcripts are tokenized with the DeBERTa vocabulary.
WavLM-Large and DeBERTa-v3-base are used with LoRA rank 16, flow dimension $d{=}64$, $\lambda_\mathrm{cls}^\mathrm{flow}{=}7$, $\sigma_{\max}{=}0.1$, $\lambda_\mathrm{back\_land}{=}0.3$, $\tau_0{=}1$, $\eta{=}0.5$, and $\tau_\mathrm{back}{=}2$ unless noted.
Within CGD-IB, $\lambda_\mathrm{cf}{=}0.5$ and $\lambda_E{=}1$; $\beta_\mathrm{IB}$ increases linearly from 0.01 to 0.1 during Stage~I and then decreases linearly to 0.02 across Stages~II.a--II.b.
Training uses AdamW (learning rate $2\times10^{-4}$, weight decay 0.01), effective batch size 32, and exponential moving average (EMA) decay 0.999.
All experiments were run on a single NVIDIA A100 GPU with 80\,GB of memory.
Unless stated otherwise, BiCFlow-MER and its controlled variants are trained using the three-stage curriculum in Table~\ref{tab:curriculum}.

\paragraph{Inference calibration.}
Using only the validation split, we fit calibrated logits $\ell_k^{\mathrm{cal}}=\ell_k/T+b_k$, where $T$ is a shared temperature and the per-class biases $b_k$ are initialized from training-set log-frequencies.
Both are selected by validation UA and frozen for testing.

\paragraph{Comparison with representative methods.}
Table~\ref{tab:external_compare} compares BiCFlow-MER with recent audio-text MER baselines, including GateM$^2$Former (GM2F)~\cite{khurana2022robinnet,zhang2023multimodal,xu2025gatem,zhang2023dual,hou2023semantic,shao2025leveraging,lei2026scmi,han2025multi}. S and T indicate speech and text access; baseline values are taken from the corresponding literature.
BiCFlow-MER ranks first on all four metrics, exceeding the second-best values by 0.40/0.11 points on IEMOCAP WA/UA and 0.55/1.43 points on MELD WA/W-F1.

\subsection{Cross-Dataset Evaluation on CASE}

BiCFlow-MER is further evaluated on Conflict in Acoustic-Semantic Emotion (CASE)~\cite{huang2026tone}, where acoustic emotion and lexical semantics may disagree.
Following the exact FAS protocol, BiCFlow-MER is trained on the same IEMOCAP, CMU-MOSEI, MER2024, MELD, RAVDESS, and ESD partitions and evaluated on the complete 378-utterance CASE set~\cite{huang2026tone}.
CASE remains strictly zero-shot, with no samples used for training, validation, calibration, or hyperparameter selection.
In the audio-only CASE configuration, we follow the FAS codebase and use its frozen Whisper-large-v2 and MingTok-Audio encoders to extract semantic- and acoustic-oriented features from each waveform.
The acoustic and semantic streams instantiate the original $a$ and $t$ branches, respectively; all downstream modules, objectives, and settings remain unchanged, and the same frozen feature extractors are used for all controlled variants.
This branch assignment does not introduce textual input: both streams are extracted from the same waveform.
Accordingly, Aco. and Sem. in Table~\ref{tab:case_results} describe the dominant information preserved by each representation rather than distinct input modalities.
No transcript encoder or textual token sequence is used in the CASE experiment.
Following the released FAS evaluation implementation, W-F1 is computed using class-support weighting.
Table~\ref{tab:case_results} reports 61.90\% accuracy (ACC) and 59.82\% W-F1, exceeding Fusion Acoustic-Semantic (FAS)~\cite{huang2026tone} by 2.52 and 4.74 points, respectively.
The remaining pretrained baselines peak at 47.26\% ACC and 44.97\% W-F1, leaving FAS and BiCFlow-MER as a distinct upper tier.
Under the strictly zero-shot protocol, these results establish end-to-end transfer to unseen acoustic-semantic conflicts; Table~\ref{tab:module_ablation} attributes this transfer to individual evidence, transport, and scoring components.

\begin{table}[!t]
\centering
\footnotesize
\setlength{\tabcolsep}{2.5pt}
\renewcommand{\arraystretch}{1.08}
\begin{tabular*}{\columnwidth}{@{}p{0.25\columnwidth}@{\extracolsep{\fill}}cccc@{}}
\toprule
Model & Aco. & Sem. & ACC(\%) & W-F1(\%) \\
\midrule
wav2vec 2.0 & \checkmark & \checkmark & 25.59 & 22.83 \\
HuBERT & \checkmark & \checkmark & 32.90 & 32.24 \\
WavLM & \checkmark & \checkmark & 34.20 & 33.92 \\
Whisper &  & \checkmark
& 47.26
& 44.97\\
CLAP &  & \checkmark & 34.46 & 31.93 \\
EnCodec & \checkmark &  & 24.54 & 18.81 \\
VibeVoice & \checkmark &  & 30.03 & 25.90 \\
MingTok-Audio & \checkmark &  & 29.24 & 25.61 \\
Emotion2Vec & \checkmark &  & 31.48 & 28.42 \\
Qwen2-Audio &  & \checkmark & 32.53 & 27.08 \\
Qwen2.5-Omni &  & \checkmark & 34.66 & 30.21 \\
FAS & \checkmark & \checkmark & \underline{59.38} & \underline{55.08} \\
\midrule
\textbf{Ours} & \checkmark & \checkmark & \textbf{\caseACC} & \textbf{\caseWF} \\
\bottomrule
\end{tabular*}
\caption{Zero-shot CASE comparison. All methods use audio-only input; Aco. and Sem. denote acoustic and semantic representation attributes. Baseline results and annotations follow the original FAS study~\cite{huang2026tone}. Best values are bolded; second-best values are underlined.}
\label{tab:case_results}
\end{table}
\subsection{Analysis and Ablation}

\begin{table}[!tb]
\centering
\footnotesize
\setlength{\tabcolsep}{2.5pt}
\renewcommand{\arraystretch}{1.08}
\begin{tabular*}{\columnwidth}{@{}l@{\extracolsep{\fill}}cccc@{}}
\toprule
\multirow{2}{*}{Variant}
& \multicolumn{2}{c}{IEMOCAP}
& \multicolumn{2}{c}{CASE} \\
\cmidrule(lr){2-3}
\cmidrule(l){4-5}
& WA(\%) & UA(\%) & ACC(\%) & W-F1(\%) \\
\midrule
\textbf{Full} & \textbf{78.18} & \textbf{78.90} & \textbf{61.90} & \textbf{59.82} \\
w/o Fact. & 75.39 & 76.54 & 56.87 & 54.30 \\
w/o ECA & 75.50 & 76.23 & 54.00 & 53.56 \\
w/o CT-BiRF & 76.22 & 76.79 & 57.01 & 56.44 \\
w/o A-GCH & 76.76 & 77.31 & 57.12 & 55.97 \\
w/o BGD & \underline{77.45} & \underline{78.19} & \underline{59.65} & \underline{58.86} \\
\bottomrule
\end{tabular*}
\caption{Controlled variants on IEMOCAP (audio-text) and zero-shot CASE (audio-only semantic/acoustic streams), with no CASE target-set adaptation. Best results are bolded; second-best results are underlined.}
\label{tab:module_ablation}
\end{table}

\begin{table}[!tb]
\centering
\small
\begin{tabular*}{\columnwidth}{@{\extracolsep{\fill}}lcccc@{}}
\toprule
\multirow{2}{*}{Method}
& \multirow{2}{*}{S}
& \multirow{2}{*}{T}
& \multicolumn{2}{c}{IEMOCAP} \\
\cmidrule(l){4-5}
& & & WA(\%) & UA(\%) \\
\midrule
Ours & \checkmark &   & 73.83$\pm$2.05 & 75.30$\pm$3.31 \\
Ours &  & \checkmark &   65.75$\pm$1.70 & 67.56$\pm$2.94 \\
Ours & \checkmark & \checkmark  & \textbf{\ourWA} & \textbf{\ourUA} \\
\bottomrule
\end{tabular*}
\caption{Modality ablation on IEMOCAP.}
\label{tab:modality}
\end{table}

\begin{figure*}[!t]
\centering
\includegraphics[width=\textwidth]{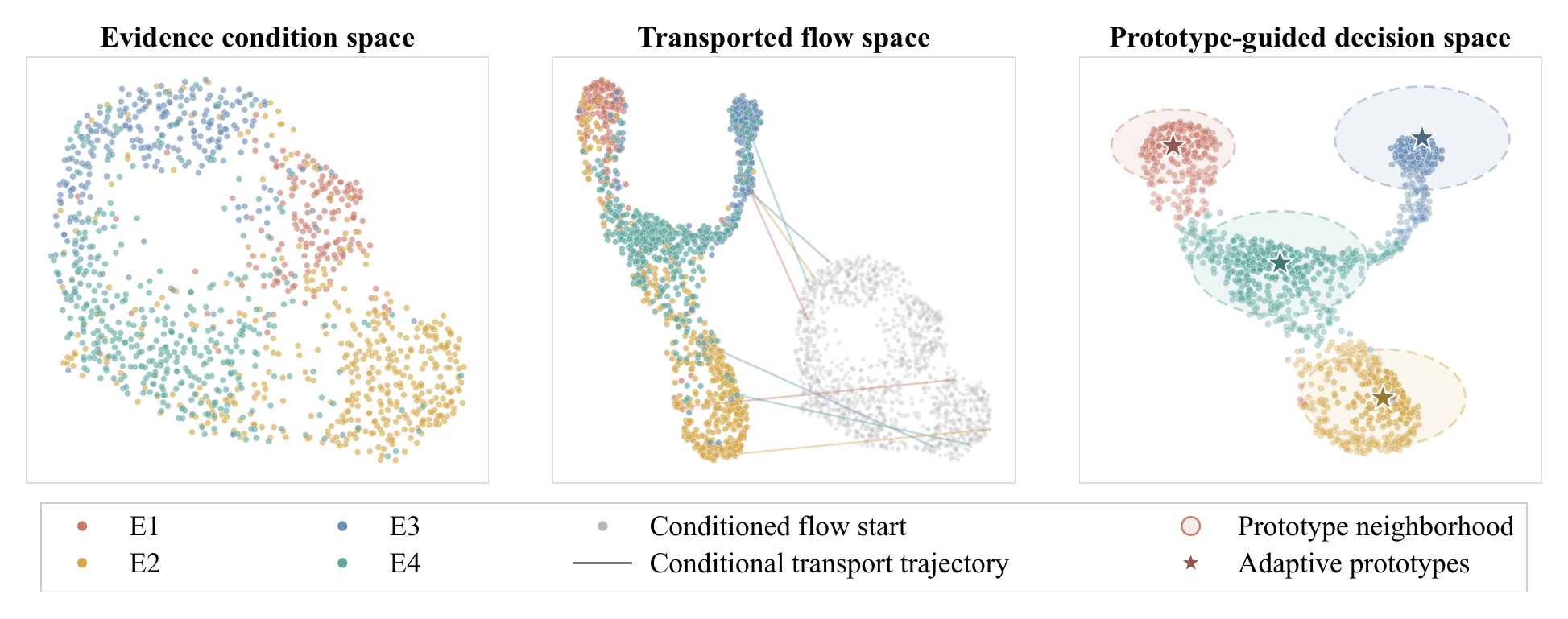}
\caption{Qualitative stage-wise organization on four-class IEMOCAP. Left: affective evidence conditions. Middle: gray condition-dependent flow starts, colored transported endpoints, and representative trajectories. Right: adaptive prototypes (stars) and schematic two-dimensional prototype neighborhoods (dashed ellipses).}
\label{fig:stage_umap}
\end{figure*}

\paragraph{Stage-wise emotion-space organization.}
E1--E4 denote angry, happy, sad, and neutral, respectively. The condition space retains local overlap, whereas CT-BiRF maps condition-dependent starts to endpoints evaluated against A-GCH prototype neighborhoods.

\begin{figure}[!t]
\centering
\includegraphics[width=\columnwidth]{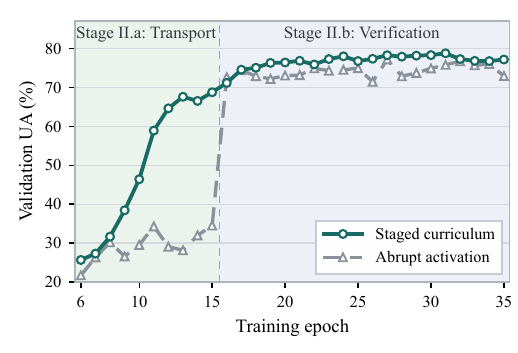}
\caption{Validation UA after Stage~I under the proposed curriculum and an abrupt-activation baseline, which activates all Stage~II components immediately after Stage~I. The proposed schedule introduces ECA, conditional transport, and geometric scoring in Stage~II.a, then adds N3 conflict pairs and backward landmark supervision in Stage~II.b.}
\label{fig:curriculum_curve}
\end{figure}

\paragraph{Curriculum behavior.}
The staged schedule reaches high validation UA earlier and remains more stable after Stage~II.b, supporting the proposed ordering.
The controlled variants below then replace each stage-specific mechanism with a simpler counterpart to test its role in this sequence.

\paragraph{Controlled variants.}
Table~\ref{tab:module_ablation} isolates the main design blocks through controlled replacements.
The w/o Fact. variant uses ordinary fused evidence; w/o ECA replaces disagreement-conditioned fusion with ordinary fusion; w/o CT-BiRF retains the same pretrained encoders but replaces conditional transport with direct recognition; w/o A-GCH removes adaptive prototype-cloud scoring; and w/o BGD removes the inference-time backward compatibility score.

On CASE, w/o ECA causes the largest degradation, with losses of 7.90 ACC and 6.26 W-F1 points; it also produces the largest IEMOCAP UA loss of 2.67 points. These results identify disagreement-conditioned fusion as an important contributor on the conflict-focused CASE benchmark.
The w/o Fact., w/o CT-BiRF, and w/o A-GCH variants reduce CASE ACC/W-F1 by 5.03/5.52, 4.89/3.38, and 4.78/3.85 points, respectively, supporting the roles of affect-oriented factorization, conditional transport, and adaptive prototype geometry in zero-shot transfer.
The w/o BGD variant produces a smaller but consistent CASE loss of 2.25 ACC and 0.96 W-F1 points, showing that backward compatibility scoring provides an additional gain.

\paragraph{Modality reliance.}
Table~\ref{tab:modality} shows that the final condition benefits from complementary speech and text evidence rather than collapsing to a single dominant stream.
Speech alone reaches 73.83$\pm$2.05 WA and 75.30$\pm$3.31 UA, while text alone reaches 65.75$\pm$1.70 WA and 67.56$\pm$2.94 UA.
The audio-text setting reaches \ourWA\% WA and \ourUA\% UA, improving on speech alone by 4.35 and 3.60 points, respectively.
CT-BiRF transports $\mathbf{e}_0$ under the affective condition $\mathbf{c}$ rather than transporting $\mathbf{c}$ itself; the bimodal gains therefore show that complementary modalities improve the condition guiding transport.

\section{Conclusion}
BiCFlow-MER addresses three evidence-level challenges in audio-text MER.
Emotion-oriented factorization separates affective evidence from speaker-style and lexical-content factors, ECA incorporates audio-text disagreement into condition construction, and conditional transport with forward geometry and backward compatibility scoring evaluates candidate emotions against both endpoint structure and the condition-dependent start region.
Results on IEMOCAP and MELD establish standard recognition performance, while the zero-shot CASE results and controlled variants identify how the complete evidence chain transfers under acoustic-semantic conflict.
This formulation establishes a generative-decision paradigm in which multimodal emotion evidence is explicitly constructed, transported, and verified.

\bibliography{arxiv-references}

@article{busso2008iemocap,
  title={IEMOCAP: Interactive emotional dyadic motion capture database},
  author={Busso, Carlos and Bulut, Murtaza and Lee, Chi-Chun and Kazemzadeh, Abe and Mower, Emily and Kim, Samuel and Chang, Jeannette N and Lee, Sungbok and Narayanan, Shrikanth S},
  journal={Language resources and evaluation},
  volume={42},
  number={4},
  pages={335--359},
  year={2008},
  publisher={Springer}
}

@inproceedings{lin2023robust,
  title={Robust multi-modal speech emotion recognition with asr error adaptation},
  author={Lin, Binghuai and Wang, Liyuan},
  booktitle={ICASSP 2023-2023 IEEE International Conference on Acoustics, Speech and Signal Processing (ICASSP)},
  pages={1--5},
  year={2023},
  organization={IEEE}
}

@inproceedings{wang2024inter,
  title={Inter-modality and intra-sample alignment for multi-modal emotion recognition},
  author={Wang, Yusong and Li, Dongyuan and Shen, Jialun},
  booktitle={ICASSP 2024-2024 IEEE International Conference on Acoustics, Speech and Signal Processing (ICASSP)},
  pages={8301--8305},
  year={2024},
  organization={IEEE}
}

@article{fan2025coordination,
  title={Coordination attention based transformers with bidirectional contrastive loss for multimodal speech emotion recognition},
  author={Fan, Weiquan and Xu, Xiangmin and Zhou, Guohua and Deng, Xiaofang and Xing, Xiaofen},
  journal={Speech Communication},
  volume={169},
  pages={103198},
  year={2025},
  publisher={Elsevier}
}

@article{heredia2022adaptive,
  title={Adaptive multimodal emotion detection architecture for social robots},
  author={Heredia, Juanpablo and Lopes-Silva, Edmundo and Cardinale, Yudith and Diaz-Amado, Jose and Dongo, Irvin and Graterol, Wilfredo and Aguilera, Ana},
  journal={Ieee Access},
  volume={10},
  pages={20727--20744},
  year={2022},
  publisher={IEEE}
}

@article{pena2023framework,
  title={A framework to evaluate fusion methods for multimodal emotion recognition},
  author={Pe{\~n}a, Diego and Aguilera, Ana and Dongo, Irvin and Heredia, Juanpablo and Cardinale, Yudith},
  journal={IEEE Access},
  volume={11},
  pages={10218--10237},
  year={2023},
  publisher={IEEE}
}

@article{liu2022flow,
  title={Flow straight and fast: Learning to generate and transfer data with rectified flow},
  author={Liu, Xingchao and Gong, Chengyue and Liu, Qiang},
  journal={arXiv preprint arXiv:2209.03003},
  year={2022}
}

@inproceedings{zhang2024camel,
  title={CAMEL: capturing metaphorical alignment with context disentangling for multimodal emotion recognition},
  author={Zhang, Linhao and Jin, Li and Xu, Guangluan and Li, Xiaoyu and Xu, Cai and Wei, Kaiwen and Liu, Nayu and Liu, Haonan},
  booktitle={Proceedings of the AAAI Conference on Artificial Intelligence},
  volume={38},
  number={8},
  pages={9341--9349},
  year={2024}
}

@inproceedings{xu2024secap,
  title={Secap: Speech emotion captioning with large language model},
  author={Xu, Yaoxun and Chen, Hangting and Yu, Jianwei and Huang, Qiaochu and Wu, Zhiyong and Zhang, Shi-Xiong and Li, Guangzhi and Luo, Yi and Gu, Rongzhi},
  booktitle={Proceedings of the AAAI Conference on Artificial Intelligence},
  volume={38},
  number={17},
  pages={19323--19331},
  year={2024}
}

@inproceedings{liu2026fine,
  title={FINE: Factorized multimodal sentiment analysis via mutual information estimation},
  author={Liu, Yadong and Wang, Shangfei},
  booktitle={Proceedings of the AAAI Conference on Artificial Intelligence},
  volume={40},
  number={3},
  pages={1955--1963},
  year={2026}
}

@inproceedings{lyu2026cross,
  title={Cross-space synergy: A unified framework for multimodal emotion recognition in conversation},
  author={Lyu, Xiaosen and Xiong, Jiayu and Chen, Yuren and Wang, Wanlong and Dai, Xiaoqing and Wang, Jing},
  booktitle={Proceedings of the AAAI Conference on Artificial Intelligence},
  volume={40},
  number={29},
  pages={24226--24234},
  year={2026}
}

@inproceedings{wu2025enriching,
  title={Enriching multimodal sentiment analysis through textual emotional descriptions of visual-audio content},
  author={Wu, Sheng and He, Dongxiao and Wang, Xiaobao and Wang, Longbiao and Dang, Jianwu},
  booktitle={Proceedings of the AAAI Conference on Artificial Intelligence},
  volume={39},
  number={2},
  pages={1601--1609},
  year={2025}
}

@inproceedings{he2026cross,
  title={Cross-modal prompting for balanced incomplete multi-modal emotion recognition},
  author={He, Wen-Jue and Zhu, Xiaofeng and Zhang, Zheng},
  booktitle={Proceedings of the AAAI Conference on Artificial Intelligence},
  volume={40},
  number={21},
  pages={17463--17471},
  year={2026}
}

@inproceedings{yin2026tical,
  title={Tical: Typicality-based consistency-aware learning for multimodal emotion recognition},
  author={Yin, Wen and Zhan, Siyu and Liu, Cencen and Hu, Xin and Duan, Guiduo and Xie, Xiurui and Li, Yuan-Fang and He, Tao},
  booktitle={Proceedings of the AAAI Conference on Artificial Intelligence},
  volume={40},
  number={21},
  pages={17948--17956},
  year={2026}
}

@inproceedings{li2026llms,
  title={Do llms feel? teaching emotion recognition with prompts, retrieval, and curriculum learning},
  author={Li, Xinran and Liu, Yu and Qiao, Jiaqi and Xu, Xiujuan},
  booktitle={Proceedings of the AAAI Conference on Artificial Intelligence},
  volume={40},
  number={38},
  pages={31778--31786},
  year={2026}
}

@inproceedings{rha2026emotion,
  title={Emotion-coherent reasoning for multimodal llms via emotional rationale verifier},
  author={Rha, Hyeongseop and Yeo, Jeong Hun and Kim, Yeonju and Ro, Yong Man},
  booktitle={Proceedings of the AAAI Conference on Artificial Intelligence},
  volume={40},
  number={3},
  pages={2029--2037},
  year={2026}
}

@article{ghosh2022mmer,
  title={Mmer: Multimodal multi-task learning for speech emotion recognition},
  author={Ghosh, Sreyan and Tyagi, Utkarsh and Ramaneswaran, S and Srivastava, Harshvardhan and Manocha, Dinesh},
  journal={arXiv preprint arXiv:2203.16794},
  year={2022}
}

@inproceedings{wang2025enhancing,
  title={Enhancing multimodal emotion recognition through multi-granularity cross-modal alignment},
  author={Wang, Xuechen and Zhao, Shiwan and Sun, Haoqin and Wang, Hui and Zhou, Jiaming and Qin, Yong},
  booktitle={ICASSP 2025-2025 IEEE International Conference on Acoustics, Speech and Signal Processing (ICASSP)},
  pages={1--5},
  year={2025},
  organization={IEEE}
}

@inproceedings{wu2023empirical,
  title={An empirical study and improvement for speech emotion recognition},
  author={Wu, Zhen and Lu, Yizhe and Dai, Xinyu},
  booktitle={ICASSP 2023-2023 IEEE International Conference on Acoustics, Speech and Signal Processing (ICASSP)},
  pages={1--5},
  year={2023},
  organization={IEEE}
}

@article{wang2026gen,
  title={Gen-SER: When the generative model meets speech emotion recognition},
  author={Wang, Taihui and Zhao, Jinzheng and Chen, Rilin and Lei, Tong and Wang, Wenwu and Yu, Dong},
  journal={arXiv preprint arXiv:2601.20573},
  year={2026}
}

@inproceedings{poria2019meld,
  title={Meld: A multimodal multi-party dataset for emotion recognition in conversations},
  author={Poria, Soujanya and Hazarika, Devamanyu and Majumder, Navonil and Naik, Gautam and Cambria, Erik and Mihalcea, Rada},
  booktitle={Proceedings of the 57th annual meeting of the association for computational linguistics},
  pages={527--536},
  year={2019}
}

@article{chen2022wavlm,
  title={Wavlm: Large-scale self-supervised pre-training for full stack speech processing},
  author={Chen, Sanyuan and Wang, Chengyi and Chen, Zhengyang and Wu, Yu and Liu, Shujie and Chen, Zhuo and Li, Jinyu and Kanda, Naoyuki and Yoshioka, Takuya and Xiao, Xiong and others},
  journal={IEEE Journal of Selected Topics in Signal Processing},
  volume={16},
  number={6},
  pages={1505--1518},
  year={2022},
  publisher={IEEE}
}

@article{he2020deberta,
  title={Deberta: Decoding-enhanced bert with disentangled attention},
  author={He, Pengcheng and Liu, Xiaodong and Gao, Jianfeng and Chen, Weizhu},
  journal={arXiv preprint arXiv:2006.03654},
  year={2020}
}

@inproceedings{cheng2020club,
  title={Club: A contrastive log-ratio upper bound of mutual information},
  author={Cheng, Pengyu and Hao, Weituo and Dai, Shuyang and Liu, Jiachang and Gan, Zhe and Carin, Lawrence},
  booktitle={International conference on machine learning},
  pages={1779--1788},
  year={2020},
  organization={PMLR}
}

@article{hu2022lora,
  title={Lora: Low-rank adaptation of large language models.},
  author={Hu, Edward J and Shen, Yelong and Wallis, Phillip and Allen-Zhu, Zeyuan and Li, Yuanzhi and Wang, Shean and Wang, Liang and Chen, Weizhu and others},
  journal={Iclr},
  volume={1},
  number={2},
  pages={3},
  year={2022}
}

@inproceedings{tsai2019multimodal,
  title={Multimodal transformer for unaligned multimodal language sequences},
  author={Tsai, Yao-Hung Hubert and Bai, Shaojie and Liang, Paul Pu and Kolter, J Zico and Morency, Louis-Philippe and Salakhutdinov, Ruslan},
  booktitle={Proceedings of the 57th annual meeting of the association for computational linguistics},
  pages={6558--6569},
  year={2019}
}

@inproceedings{yu2021learning,
  title={Learning modality-specific representations with self-supervised multi-task learning for multimodal sentiment analysis},
  author={Yu, Wenmeng and Xu, Hua and Yuan, Ziqi and Wu, Jiele},
  booktitle={Proceedings of the AAAI conference on artificial intelligence},
  volume={35},
  number={12},
  pages={10790--10797},
  year={2021}
}

@inproceedings{schroff2015facenet,
  title={Facenet: A unified embedding for face recognition and clustering},
  author={Schroff, Florian and Kalenichenko, Dmitry and Philbin, James},
  booktitle={Proceedings of the IEEE conference on computer vision and pattern recognition},
  pages={815--823},
  year={2015}
}

@inproceedings{peebles2023scalable,
  title={Scalable diffusion models with transformers},
  author={Peebles, William and Xie, Saining},
  booktitle={Proceedings of the IEEE/CVF international conference on computer vision},
  pages={4195--4205},
  year={2023}
}

@article{kingma2013auto,
  title={Auto-encoding variational bayes},
  author={Kingma, Diederik P and Welling, Max},
  journal={arXiv preprint arXiv:1312.6114},
  year={2013}
}

@article{wu2026amber,
  title={{AmbER}\textsuperscript{2}: Dual Ambiguity-Aware Emotion Recognition Applied to Speech and Text},
  author={Wu, Jingyao and Lin, Grace and Song, Yinuo and Picard, Rosalind},
  journal={arXiv preprint arXiv:2601.18010},
  year={2026}
}

@article{khurana2022robinnet,
  title={RobinNet: A multimodal speech emotion recognition system with speaker recognition for social interactions},
  author={Khurana, Yash and Gupta, Swamita and Sathyaraj, R and Raja, SP},
  journal={IEEE Transactions on Computational Social Systems},
  volume={11},
  number={1},
  pages={478--487},
  year={2022},
  publisher={IEEE}
}

@article{zhang2023multimodal,
  title={Multimodal emotion recognition based on audio and text by using hybrid attention networks},
  author={Zhang, Shiqing and Yang, Yijiao and Chen, Chen and Liu, Ruixin and Tao, Xin and Guo, Wenping and Xu, Yicheng and Zhao, Xiaoming},
  journal={Biomedical Signal Processing and Control},
  volume={85},
  pages={105052},
  year={2023},
  publisher={Elsevier}
}

@inproceedings{xu2025gatem,
  title={Gatem 2 former: Gated feature selection and expert modeling in multimodal emotion recognition},
  author={Xu, Weixiang and Dong, Zhongren and Wang, Runming and Xu, Xinzhou and Zhang, Zixing},
  booktitle={ICASSP 2025-2025 IEEE International Conference on Acoustics, Speech and Signal Processing (ICASSP)},
  pages={1--5},
  year={2025},
  organization={IEEE}
}

@inproceedings{zhang2023dual,
  title={A Dual Attention-based Modality-Collaborative Fusion Network for Emotion Recognition.},
  author={Zhang, Xiaoheng and Li, Yang},
  booktitle={INTERSPEECH},
  pages={1468--1472},
  year={2023}
}

@article{hou2023semantic,
  title={Semantic alignment network for multi-modal emotion recognition},
  author={Hou, Mixiao and Zhang, Zheng and Liu, Chang and Lu, Guangming},
  journal={IEEE Transactions on Circuits and Systems for Video Technology},
  volume={33},
  number={9},
  pages={5318--5329},
  year={2023},
  publisher={IEEE}
}

@inproceedings{shao2025leveraging,
  title={Leveraging Label Potential for Enhanced Multimodal Emotion Recognition},
  author={Shao, Xuechun and Yu, Yinfeng and Wang, Liejun},
  booktitle={2025 International Joint Conference on Neural Networks (IJCNN)},
  pages={1--8},
  year={2025},
  organization={IEEE}
}

@article{lei2026scmi,
  title={SCMI-Net: Semantic constraints and modal interaction network for multimodal emotion recognition},
  author={Lei, Jianjun and Zhang, Tao and Wang, Ying and Wang, Yumei},
  journal={Journal of Intelligent Information Systems},
  pages={1--23},
  year={2026},
  publisher={Springer}
}

@article{han2025multi,
  title={Multi-Level Interaction for Emotion Recognition from Unaligned Speech and Text},
  author={Han, Lingmin and Chen, Xianhong and Jia, Maoshen and Bao, Changchun},
  journal={IEEE Transactions on Affective Computing},
  year={2025},
  publisher={IEEE}
}

@article{huang2026tone,
  title={When Tone and Words Disagree: Towards Robust Speech Emotion Recognition under Acoustic-Semantic Conflict},
  author={Huang, Dawei and Lv, Yongjie and Xiong, Ruijie and Jin, Chunxiang and Peng, Xiaojiang},
  journal={arXiv preprint arXiv:2601.04564},
  year={2026}
}

\end{document}